\documentclass[letterpaper, 10 pt, conference]{ieeeconf}  

\usepackage{graphicx}
\usepackage{amsmath}
\usepackage{amssymb}
\usepackage{booktabs}
\usepackage{algorithm}
\usepackage{algpseudocode}
\usepackage{float}
\usepackage{hyperref}
\usepackage{multirow}

\IEEEoverridecommandlockouts                              

\title{\LARGE \bf
Roundabout Dilemma Zone Data Mining and Forecasting with Trajectory Prediction and Graph Neural Networks
}

\author{Manthan Chelenahalli Satish$^{1}$, Duo Lu$^{2}$, Bharatesh Chakravarthi$^{1}$, Mohammad Farhadi$^{3}$ and Yezhou Yang$^{1}$
\thanks{$^{1}$M. Satish, B. Chakravarthi, and Y. Yang are with Arizona State University.
{\tt\small \{mcsatish, bshettah, yz.yang\}@asu.edu}}%
\thanks{$^{2}$D. Lu is with Rider University.
{\tt\small \{dlu\}@rider.edu}}%
\thanks{$^{3}$M. Farhadi is with ARGOS Vision Inc.
{\tt\small \{farhadi\}@argos. vision}}%
\thanks{This research is sponsored by NSF grant \#2038666, \#2329780, the Institute of Automated Mobility (Arizona), and Rider University. We thank Arizona DOT, Greg Leeming, and Marisa Paula Walker for their help.}
\thanks{Dataset, visualization, and more qualitative results will be made publicly available at \href{https://github.com/mnthnx64/roundabout-dilemma-zone}{github.com/mnthnx64/roundabout-dilemma-zone}.}%
}

\begin{document}

\maketitle

\thispagestyle{empty}
\pagestyle{empty}

\begin{abstract}
Traffic roundabouts, as complex and critical road scenarios, pose significant safety challenges for autonomous vehicles. In particular, the encounter of a vehicle with a dilemma zone (DZ) at a roundabout intersection is a pivotal concern. This paper presents an automated system that leverages trajectory forecasting to predict DZ events, specifically at traffic roundabouts. Our system aims to enhance safety standards in both autonomous and manual transportation. The core of our approach is a modular, graph-structured recurrent model that forecasts the trajectories of diverse agents, taking into account agent dynamics and integrating heterogeneous data, such as semantic maps. This model, based on graph neural networks, aids in predicting DZ events and enhances traffic management decision-making. We evaluated our system using a real-world dataset of traffic roundabout intersections. Our experimental results demonstrate that our dilemma forecasting system achieves a high precision with a low false positive rate of 0.1. This research represents an advancement in roundabout DZ data mining and forecasting, contributing to the assurance of intersection safety in the era of autonomous vehicles.

\end{abstract}

\section{INTRODUCTION}

\label{sec:intro}
Drivers often face dilemmas at traffic intersections, where they must decide whether to stop or proceed through the intersection \cite{zhang_fu_hu_2014}\cite{Papaioannou2021DilemmaZM}. 
Yellow-light dilemma zone (DZ) is the area on a roadway where a driver may encounter such a dilemma.
This dilemma can lead to dangerous situations, such as red-light running (RLR) or rear-ended crashes, particularly at high-volume or high-speed intersections \cite{zhang_fu_hu_2014}. Various studies have investigated the factors affecting drivers' decision-making processes, such as gap size, speed, and visibility \cite{zhang_fu_hu_2014}\cite{gates_2018}\cite{li_tan_lin_2021}.

\begin{figure}[t]
  \centering
    \includegraphics[width=0.95\linewidth]{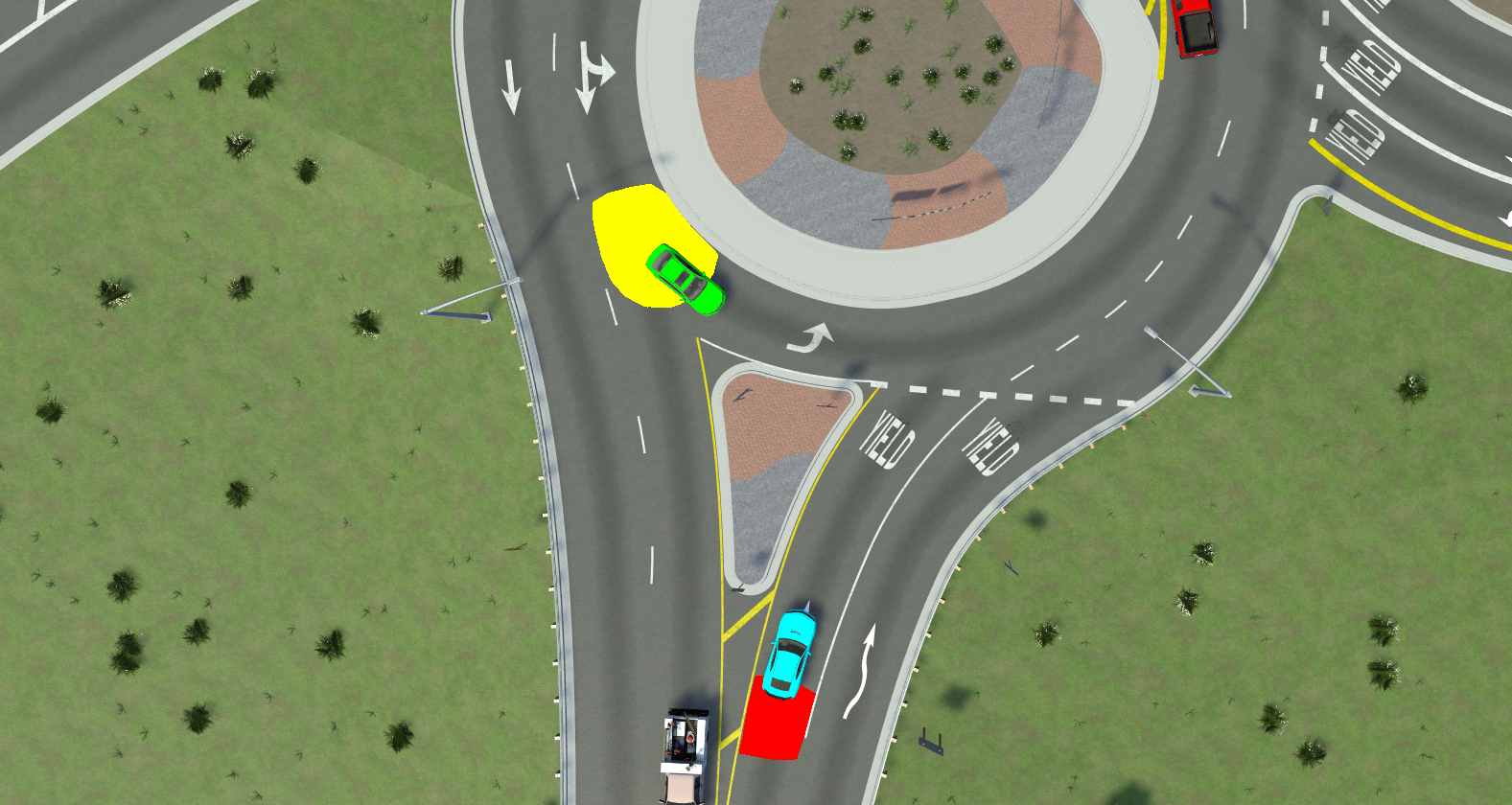}
    \caption{An illustration of a roundabout dilemma zone event. Here, a car in the yellow zone is insinuating a ``moving yellow light", creating a dilemma zone that influences the car in the red zone entering the roundabout.
    }
    \label{fig:rd_dz}
    \vspace{-0.2in}
\end{figure}

Transportation engineers have proposed various solutions, such as advanced warning signs or pavement markings and intelligent transportation systems (ITS) using pressure, radar, or camera-based technologies to detect and monitor yellow-light DZs \cite{wu_ma_li_2013}\cite{park_chang_2017}\cite{brasil_machado_2017}\cite{nguyen}. However, these solutions are not applicable to roundabouts that do not have traffic lights.
Instead, roundabouts use yield-controlled entries and continuous circular flows to improve traffic flow efficiency, reduce delays, and enhance safety. As a result, roundabouts can also create DZs where drivers may struggle to judge the safe gap size or time to enter the roundabout since vehicles inside the roundabouts create conflicts similar to ``moving yellow lights", as shown in Fig. \ref{fig:rd_dz}. In recent years, roundabout DZs have gained attention due to their unique characteristics and challenges \cite{wei_li_gong_gong_li_2021}, such as difficulties in obtaining accurate training data for driver behavior prediction \cite{li_tan_lin_2021}\cite{xiong_narayanaswamy_bao_flannagan_sayer_2016}, constantly changing factors affecting driver decision-making \cite{zhang_fu_hu_2014}\cite{li_jia_shao_2016}, and tracking errors from roadside units \cite{lu2021carom}.

\begin{figure*}[!h]
  \centering
  \includegraphics[width=0.95\linewidth]{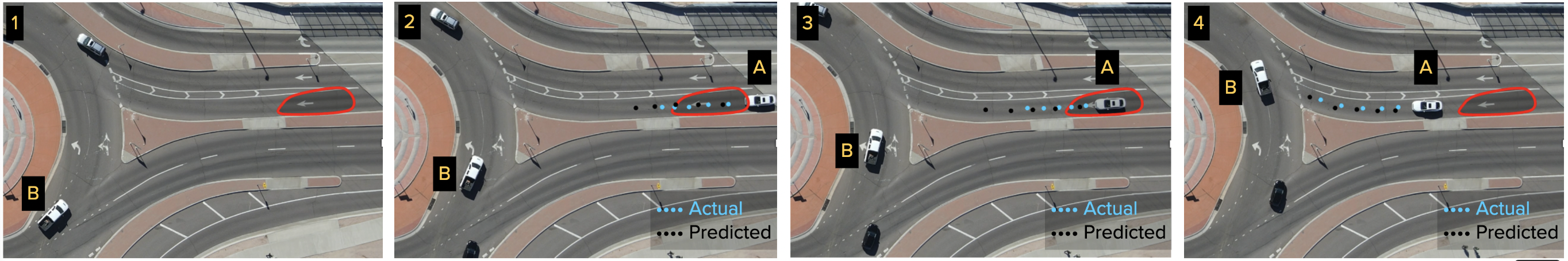}
   \caption{
   An example of the driving behavior in a roundabout DZ: (1) Vehicle `B' inside the roundabout triggers a DZ as the area circled in red. (2) Vehicle `A' enters the DZ, and its trajectory is predicted. (3) Vehicle `A' rapidly decelerates, causing extensive deviation between the predicted trajectory and the actual trajectory. (4) Upon exit, vehicle `A' resumes, and its predicted trajectory agrees with the actual trajectory. 
   } 
   \label{fig:teaser}
   \vspace{-0.2in}
\end{figure*}

To tackle the challenges associated with roundabout DZ, we propose a practical data mining approach to detect abnormal driving behaviors and filter out roundabout DZ events given the road geometry and vehicle trajectories. Our method compares the predicted trajectory with the actual trajectory, as shown in Fig. \ref{fig:teaser}. Specifically, our trajectory prediction model incorporates multiple pieces of information to make its prediction, such as the driver's speed and acceleration, the position of other vehicles, and a semantic map of the environment. If there is a discrepancy between the predicted trajectory and the actual trajectory, it means that the driver has had to brake hard, accelerate rapidly, or make a sudden turn, which is indicative of the vehicle being in a DZ \cite{li_jia_shao_2016}. Further, using a graph representation of the vehicles, we extend the trajectory predictor to forecast DZ events on the fly from the perspective of an autonomous vehicle, which can help the vehicle to make a decision.

We also utilize accurate vehicle tracking through drone data \cite{duolu_2022}, which provides a unique bird's eye view (BEV) perspective \cite{bev} that is not easily obtainable by other means. However, it is important to note that the use of drone data is not universal, and there are limitations to collecting data for extended periods. However, in the context of autonomous vehicles, the BEV representation can also be generated using data from the vehicles themselves \cite{liu2022bevfusion}. Modern autonomous vehicles are typically equipped with advanced sensors, such as LiDAR, radar, and cameras, which can be used to create a comprehensive view of the vehicle's surroundings. By fusing data from these sensors, it is possible to generate a BEV perspective that is similar to the one provided by drones.

In summary, our contributions are as follows.

\begin{itemize}
\item We introduce a data mining method to effectively identify roundabout DZ occurrences using vehicle trajectories. Our method leverages trajectory prediction and abnormal driving behavior patterns in the DZ.
\item We propose a dilemma forecasting method with a graph neural network that estimates the likelihood that a given vehicle encounters a DZ at every moment. It can also suggest proactive measures to ensure safety.
\item We demonstrate the effectiveness of our method through extensive experiments on a dataset of real-world roundabouts, providing insights into the practicality.
\end{itemize}

\section{RELATED WORKS}
\subsection{Roundabout Dilemma Zone}

The roundabout DZ \cite{Papaioannou2021DilemmaZM}\cite{li_jia_shao_2016}\cite{najmi_choupani_aghayan_2019} is a brief period when a vehicle must decide whether to yield or pass when approaching a roundabout, based on the presence of other vehicles inside the roundabout. This can be a tricky decision for drivers, as it requires good judgment of the speed and distance of other vehicles. Improper judgments can result in traffic violations and sometimes even rear-ended crashes \cite{brasil_machado_2017}. Decision-making can be challenging, particularly with large traffic volumes or when traffic is moving at high speeds.

\begin{figure}[!h]
  \centering
	\includegraphics[width=0.95\linewidth]{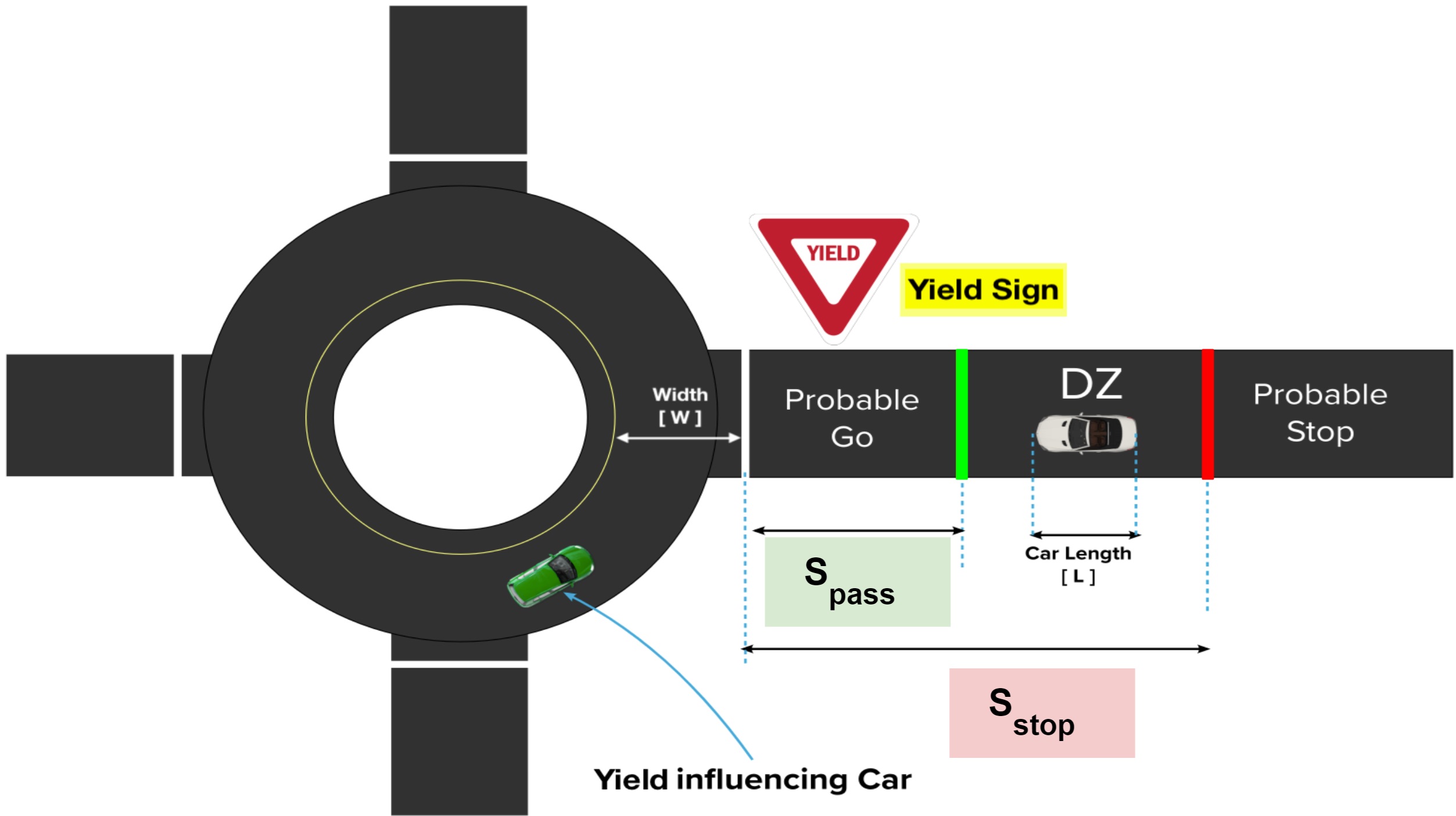}
	
   \caption{Formulation of the roundabout DZ.}
   \label{fig:concept}
   \vspace{-0.1in}
\end{figure}

Fig. \ref{fig:concept} shows a vehicle in a dilemma zone. The mathematical model \cite{zhang_fu_hu_2014}\cite{Papaioannou2021DilemmaZM} of a vehicle of length $L$ approaches a roundabout yield scenario \cite{usdot_signal}. If the driver's reaction time is given by $\delta_{react}$ and the vehicle is traveling at an initial velocity $v_0$ and accelerating at $a_{acc}$ m/s, then the maximum distance at which the vehicle can pass safely is given by:
\begin{equation} 
\label{eqn:somelabel2}
	S_{pass} = W + L + \frac{1}{2}a_{acc} \delta_{react}^2 .
\end{equation}
Where $W$ is the width of the road in the intersection.

Similarly, if the vehicle has a safe braking deceleration of $a_{dec}$, the minimum distance before at which the vehicle needs to stop is given by,

\begin{equation} 
\label{eqn:somelabel}
	S_{stop} = v_0 \delta_{react} + \frac{v_0^2}{2a_{dec}} .
\end{equation}

A vehicle is said to be in a dilemma if it lies between the distances  $S_{stop} - S_{pass}$ at a given yield signal. The case when $S_{stop} > S_{pass}$ is a DZ, whereas the case when $S_{stop} < S_{pass}$ is called an option zone.

The behavior of vehicles in yellow-light DZs at intersections has been widely studied \cite{xiong_narayanaswamy_bao_flannagan_sayer_2016}\cite{li_jia_shao_2016}\cite{yan_radwan_klee_guo_2005} and several factors were brought to light on the cause and effect. The three major factors to be considered are (a) approaching velocity, (b) acceleration after the event starts, and (c) the type of the vehicle incurring the yield. In the study conducted by li e. al. \cite{li_jia_shao_2016} for yellow light DZs at signalized intersections, 26\% of their drivers chose to stop, whereas 74\% of their drivers were observed to run through the intersection. After the yellow signal started, the majority of vehicles elected to avoid the intersection, stopping only when there was a strong likelihood that they would commit a Red Light Running (RLR) violation. In other simulated studies \cite{yan_radwan_klee_guo_2005}, it was noticed that drivers stopping when in a DZ decelerate rapidly, which could contribute to rear-ended clashes.

While some studies have examined dilemmas at roundabouts, such as the one conducted by Najmi et al. \cite{najmi_choupani_aghayan_2019}, they do not fully extend the concept of normal DZs to roundabout DZs due to the significant presence of deceleration and priority given to circular traffic flow. The study also found that the dilemma zone in roundabouts was shorter in length and closer to the stop line. 
We extend the studies presented in existing works\cite{xiong_narayanaswamy_bao_flannagan_sayer_2016}\cite{li_jia_shao_2016}\cite{yan_radwan_klee_guo_2005} to roundabouts by understanding the behavior of vehicles and the effect of the yield sign on them. We particularly solve this issue by generalizing a roundabout as an intersection with virtual signals and by predicting the trajectories of the vehicles present in the scene. This system can help drivers make informed decisions when approaching a roundabout, reducing the risk of accidents and improving traffic flow.

\subsection{Trajectory Prediction}
Social connections and spatial environment \cite{gupta_johnson_fei_2018}\cite{liang_jiang_hauptmann_2020} can affect how people and vehicles will move in the future. In the case of the DZ, the behavior of vehicles is characterized by vehicle dynamics. Works that consider dynamics \cite{salzmann_ivanovic_chakravarty_pavone_2020}\cite{li_yang_tomizuka_choi_2020} use a graph-based approach to model the agent environment. This involves adding features of the agent as the features of a node in a graph. Salzmann et al. \cite{salzmann_ivanovic_chakravarty_pavone_2020} proposed Trajectron++, an approach that involves heterogeneous data such as semantic maps of the agent environment to predict its trajectory by using a convolutional neural network (CNN).
Acceleration and lane information highly characterize DZ conditions. These features can be leveraged to detect abnormalities in normal driving behavior. However, features such as acceleration and lane information that affect a vehicle's behavior in DZ can be added to refine its predictions.

\section{Roundabout Dilemma Zone Data Mining}

\subsection{Dilemma Zone Determination}

To unearth DZ data concealed within datasets of vehicle trajectories, we employ the equations Eq. \ref{eqn:somelabel2} and \ref{eqn:somelabel} designed to estimate the boundaries of the dilemma zone on the road surface. By plugging in the speed limits of the intersection, the safe acceleration, and the deceleration rate into this equation, we can calculate an approximate region where DZ events are likely to occur.

One pivotal aspect of our data mining strategy involves when to apply the calculation with the two equations. Compared to yellow-light DZ at signalized intersections, we developed a procedure to determine when a vehicle inside the roundabout generates a ``virtual yellow light" for vehicles approaching the roundabout, which is shown in Algorithm \ref{alg:correlation}. It is based on driving safety metrics \cite{wishart}\cite{minderhoud_bovy_2001} such as time to collision (TTC) and time to stop (TTS). There are two key observations as follows.
\begin{itemize}
    \item \textit{Identification of Virtual Signals:} A vehicle inside the roundabout requires vehicles entering the roundabout to yield, which is similar to a yellow light or a red light at intersections. A yellow light indicates slowing down and proceeding with caution, while a red light indicates stopping and waiting due to trajectory conflict.
    \item \textit{Assumption of True Dilemma:} Any vehicle exhibiting a yellow light signal within the vicinity of the calculated DZ is considered to be a true DZ event. The proximity of these vehicles to the DZ boundary, within a distance denoted as $D_t$ signifies their involvement in the event.
\end{itemize}
We use this DZ determination method to filter out ground-truth DZ events to train our abnormal driving event detector detailed in the next subsection.

\begin{algorithm}[t]
\caption{Virtual Yellow Light Generation}
\begin{algorithmic}[1]
\Procedure{CalculateTrafficSign}{}
\State $Signal \gets \text{Green}$
\For{$car_i$ approaching the roundabout}
\State $pos_i \gets \text{Posistion($car_i$)}$
\For{$car_j$ in the roundabout}
\State $pos_j \gets \text{Position($car_j$)}$
\State $TTC \gets \text{TimeToCollision($car_i$, $car_j$)}$
\State $TTS \gets \text{TimeToStop($car_i$)}$
\State $SOC \gets \text{SeparationOf($car_i$, $car_j$)}$
\State $t \gets \text{Current time}$
\State $T \gets \text{Time of entry of $car_i$ into roundabout}$
\State $T_{\text{max}} \gets \text{Time limit for safe entry}$
\State $D_t \gets \text{Distance limit for influence}$
\If{$\big(TTC<T_{\text{max}} \big) \wedge \big( SOC < D_t \big)$}
\If{$\big(TTC < TTS \big) $}
\State $Signal \gets \text{Red}$
\State \textbf{return} Signal
\ElsIf{$\big(TTC > TTS \big)$}
\State $Signal \gets \text{Yellow}$
\EndIf
\EndIf
\EndFor
\State \textbf{return} Signal
\EndFor
\EndProcedure
\end{algorithmic}
\label{alg:correlation}

\end{algorithm}

\subsection{Abnormal Driving Event Detection}
Predicting the trajectory of a vehicle on the road would serve as the ground for detecting dilemma events. Salzmann et. al \cite{salzmann_ivanovic_chakravarty_pavone_2020} proposed a forecasting model that was capable of incorporating agent dynamics with semantic maps (heterogeneous data) in a multi-agent scenario. 
We propose Superpowered Trajectron++ (ST++), which is a modified version of Trajectron++ \cite{salzmann_ivanovic_chakravarty_pavone_2020} trained with the features that have a significant impact \cite{Papaioannou2021DilemmaZM}\cite{li_jia_shao_2016} on the dilemma zone to predict a vehicle's trajectory under non-dilemma conditions.
Under a dilemma zone, the behavior of the driver changes drastically, which is abnormal. 
To leverage this knowledge, a trajectory prediction model that produces a target probability $p(\mathbf{y | x})$ where $\mathbf{x} = \mathbf{s}_{1,...,N(t)}^{t-H:t} \in \mathbb{R}^{(H+1)\times N(t) \times D}$ are the current and previous D-dimensional states of the modeled agents is trained under non-dilemma conditions. The model outputs the probability distribution of the output trajectory $p(\mathbf{y} \mid \mathbf{x})=\sum_{z \in Z} p_\psi(\mathbf{y} \mid \mathbf{x}, z) p_\theta(z \mid \mathbf{x})$ where $z \in Z$ is a discrete Categorical latent variable that encodes high-level latent behavior, $\psi$ and $\theta$ are deep neural network weights that parameterize their respective distributions. The most likely trajectory $\mathbf{y}_{pred}$ can be calculated from the above distribution:
$$
z_{\text {mode }}=\arg \max _z p_\theta(z \mid \mathbf{x}),
$$
\begin{equation}
    \label{eqn:mostlikely}
    \mathbf{y}_{pred}=\arg \max _{\mathbf{y}} p_\psi\left(\mathbf{y} \mid \mathbf{x}, z_{\text{mode}}\right).
\end{equation}
As a driver behaves abnormally under a DZ, we can compute the deviation of the driver's trajectory and the model's predicted trajectory:
\begin{equation} 
    \label{eqn:deviation}
        \Delta Path = \sum_{i=1}^n ||\mathbf{y}_{pred} - \mathbf{y}_{true}||_2 .
\end{equation}

However, a dilemma cannot the calculated just by thresholding $\Delta Path$.
The driver would have made an abrupt lane change, slowed down, or accelerated on purpose, which would deviate from the model's predicted trajectory. An expert classifier is required to classify if $\Delta Path$ belongs to a DZ event or not. In our detector, we use a shallow neural network that takes into input the path deviations of the future four timesteps $\Delta_1,... ,\Delta_4$ whose details are given in Table \ref{tab:network_architecture}. Other information can also be combined in future work.

\begin{table}[H]
  \centering
  \begin{tabular}{@{}l|cccc@{}}
    \toprule
    \textbf{Layer} & \textbf{Input} & \textbf{Output} & \textbf{Activation} & \textbf{Size} \\
    \midrule
   	Input & $\Delta_1, ... ,\Delta_4$ & $\mathbf{x}$ & None & $4\times 32$ \\
   	Hidden & $\mathbf{x}$ & $\mathbf{g}$ & Sigmoid & $32 \times 1$ \\
   	Output & $\mathbf{g}$ & $\mathbf{z}$ & Sigmoid & $1 \times 1$ \\
    \bottomrule
  \end{tabular}
  \caption{Network architecture for dilemma detection}
\label{tab:network_architecture}
\vspace{-0.2in}
\end{table}

Finally, the abnormal driving prediction is made by thresholding $\mathbf{z}$ to be 0.5 after training.

\section{Dilemma Forecasting and Decision Making}

Our forecasting method leverages the state graph that was originally employed in the ST++ trajectory prediction algorithm. This state graph serves as a foundational framework for modeling vehicle movements and interactions within a roundabout.
Within this, we implement a graph neural network node classifier designed to assess the likelihood of a node (or a vehicle) ending up in a dilemma situation in the future ($P_{Dilemma}$). We also calculate the probability of a node causing a dilemma $P_{Causal}$. Calculating the probability of a node causing a dilemma is important because it helps identify vehicles that are likely to create a dilemma zone. By understanding which vehicles have a higher probability of causing a dilemma, we can take proactive measures to prevent such situations from occurring. This classification process hinges on the analysis of various parameters and historical data, allowing us to make informed predictions about impending DZ scenarios.

\begin{figure}[h]
  \centering
    \includegraphics[width=0.80\linewidth]{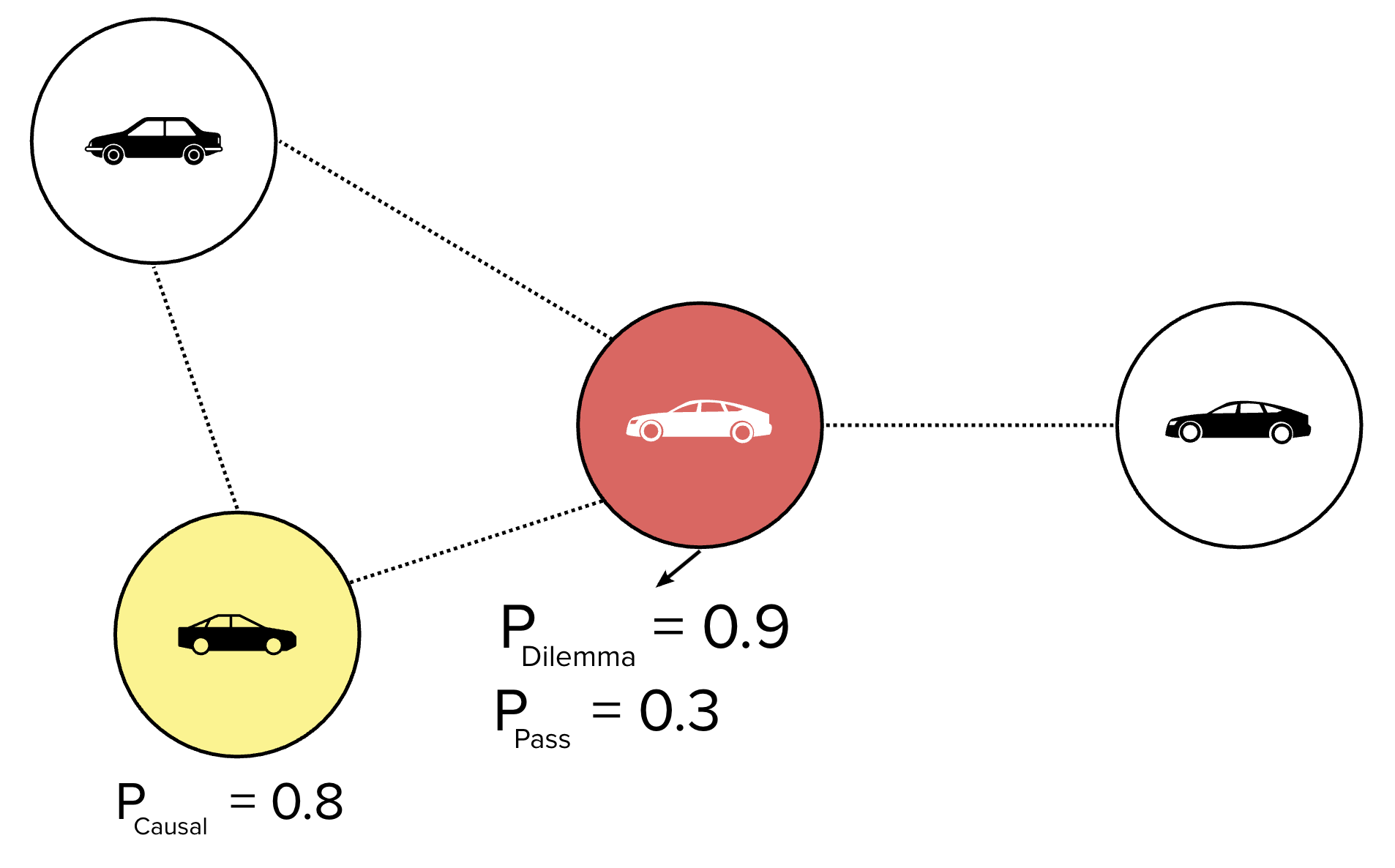}
    \caption{Dilemma forecasting with graph neural network.}
  \label{fig:superpowered}
\end{figure}

To train our dilemma event forecasting model, we harness the wealth of data generated by ST++. The training process involves feeding the trajectories with the information of them falling into a DZ into our forecasting model and fine-tuning the model's parameters. This iterative process refines the model's ability to identify potential DZ events based on real-world trajectories and empirical data.

\section{Experimental Evaluation}

\subsection{Experiment Setup}

\textbf{Dataset.} We evaluated our methods using the publicly available CAROM Air dataset\footnote{https://github.com/duolu/CAROM} curated by Lu et al. \cite{duolu_2022}. This dataset is a collection of videos of vehicle trajectories in the United States recorded by a drone. There are diverse sites covered in the dataset, such as roundabouts, intersections, local road segments, and highways. The dataset has a variety of features, including the local position coordinates of vehicles, their velocity and acceleration, and semantic maps of the intersections. Each site has approximately 60 minutes of footage, and there are approximately 4-5 vehicles in each frame. We use the vehicle trajectory data of ten roundabout sites in the dataset.

To calculate dilemma zones on the road surface, we use the speed limits suggested by the U.S. Department of Transportation \cite{usdot}. The speed is constrained between 15-25 Mph for any oncoming vehicle, irrespective of the roundabout and regardless of the posted speed limits on approaches. 

We calculate the dilemma zone by using equations \ref{eqn:somelabel2} and \ref{eqn:somelabel}. We plug in the speed limit of the intersection as the velocity of the vehicle and the safe deceleration or braking rate \cite{usdot_signal} as $a_{dec}$. 

\textbf{Baselines.} To compare the modified trajectory prediction model, we use the original Trajectron++ as the baseline, which is trained without DZ features on the same dataset.

\textbf{Metrics.} For dilemma zone event detection, Two sets of metrics are used as follows. 
\begin{enumerate}
    \item False Positive Rate (FPR) is used as the main metric to evaluate the dilemma zones detected by the system. This is used as the primary metric for DZ classification as the model would face a high number of non-dilemma scenarios, as dilemma zone events are rare. Additionally, the F1 score and recall are used to show the overall performance of the detector.
    \item Intersection over Union (IoU) measures the overlap between the detected dilemma zones and the ground truth dilemma zones. 
\end{enumerate}

For trajectory forecasting, two metrics are used as follows.
\begin{enumerate}
    \item Average Displacement Error (ADE), which is the average of L2 norms between the $i^{th}$ position of the ground truth and the prediction horizon.
    \item Final Displacement Error (FDE), which is the L2 norm between the predicted final position and the ground truth final position at the prediction horizon.
\end{enumerate}

\subsection{Implementation Details}

We implement the system with Python and PyTorch \cite{paszke2019pytorch}. Using the dataset, we demonstrate the trajectory prediction model and detection of dilemma zones on Roundabout-A5 as the number of dilemma events is large.
80\% of the video is randomly sampled and used for training, whereas test and validation sets are comprised of 10\% each. Before training, the dilemma zones are calculated from the training video by exhaustively checking each vehicle pair using Algorithm \ref{alg:correlation}. The ST++ is trained for 100 epochs with a batch size of 64. The DZ classification network is trained for 100 epochs with a batch size of 128. All the models are trained on a single Nvidia GeForce GTX 1080 GPU. The hyperparameters of the model are set to $H = 3$, Time of entry $T = 4, T_{\text{max}} = 1.5s, D_t = 10m, \alpha = 0.5, \gamma = 0.2$ where $H$ and $T$ are the number of timesteps in the input trajectory and the number of timesteps in the prediction window respectively. We threshold $P_{Pass}$ and $P_{Dilemma}$ at 0.5 for dilemma forecasting.

\begin{figure}[t]
  \centering
	\includegraphics[width=0.85\linewidth]{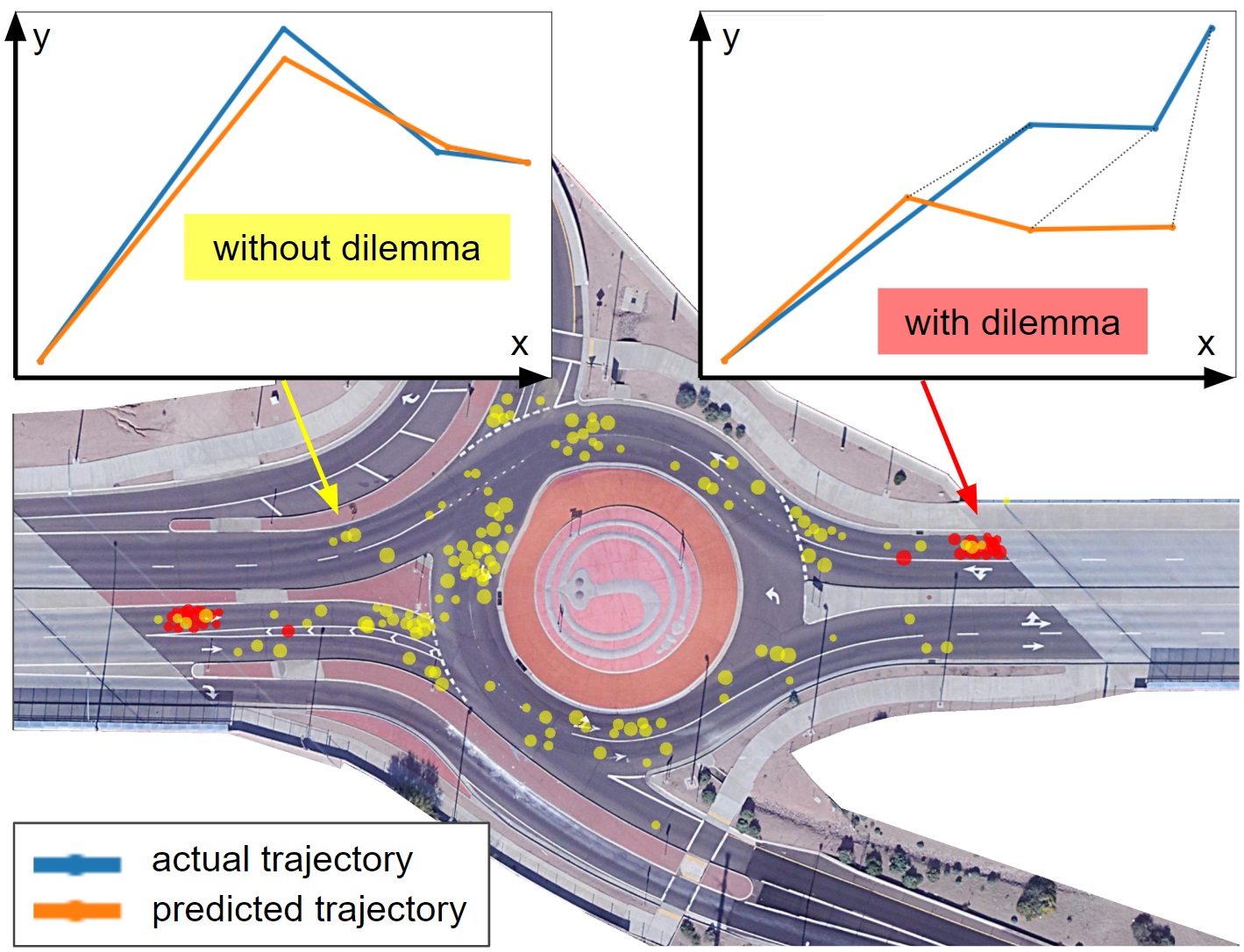}
   \caption{Abnormal driving events detected using trajectory deviation from the prediction with dilemma (red) and without dilemma (yellow).}
   \label{fig:trajectory_dev_2}
  \vspace{-0.2in}
\end{figure}

To evaluate the effectiveness of our proposed method, we observe the trajectory deviation of a vehicle in a DZ versus that of a non-DZ, as shown in Figure \ref{fig:trajectory_dev_2}. 
The trajectory deviation is defined as the ratio of the L2 norm of the difference between the predicted trajectory and the ground truth trajectory to the L2 norm of the ground truth trajectory.
All the path deviations that had a deviation larger than 0.8 (presence of a large deviation) at one of the predicted timesteps were collected. But, due to the presence of only a small portion of the deviations that belonged to the DZ  (20\% of Non-DZ), we do account for an equal portion of the largest deviations in the non-dilemma zone. This was used as the dataset to train the shallow network listed in Table \ref{tab:network_architecture}. The results are shown in Table \ref{tab:classification_results}

\begin{table}[h]
  \centering
  \begin{tabular}{@{}l|l|cccc@{}}
    \toprule
    \textbf{Type} & \textbf{Data} &  \textbf{F1 score} & \textbf{Recall}& \textbf{FPR} & \textbf{IoU} \\
    \midrule
    DZ Detection    & Training      & 0.93  & 0.92  & 0.14  & 0.92 \\
                    & Validation    & 0.92  & 0.93  & 0.06  & 0.89 \\
   	            & Testing       & 0.91  & 0.95  & 0.10  & 0.87 \\
    DZ Forecasting  & Training      & 0.91  & 0.94  & 0.09  &  -   \\
                    & Validation    & 0.88  & 0.90  & 0.14  &  -   \\
                    & Testing       & 0.83  & 0.81  & 0.17  &  -   \\
    \bottomrule
  \end{tabular}
  \caption{ Evaluation of Dilemma forecasting system. }
  \label{tab:classification_results}
  \vspace{-0.3in}
\end{table}

\subsection{Ablation Study}
\begin{figure}[t]
  \centering
	\includegraphics[width=0.95\linewidth]{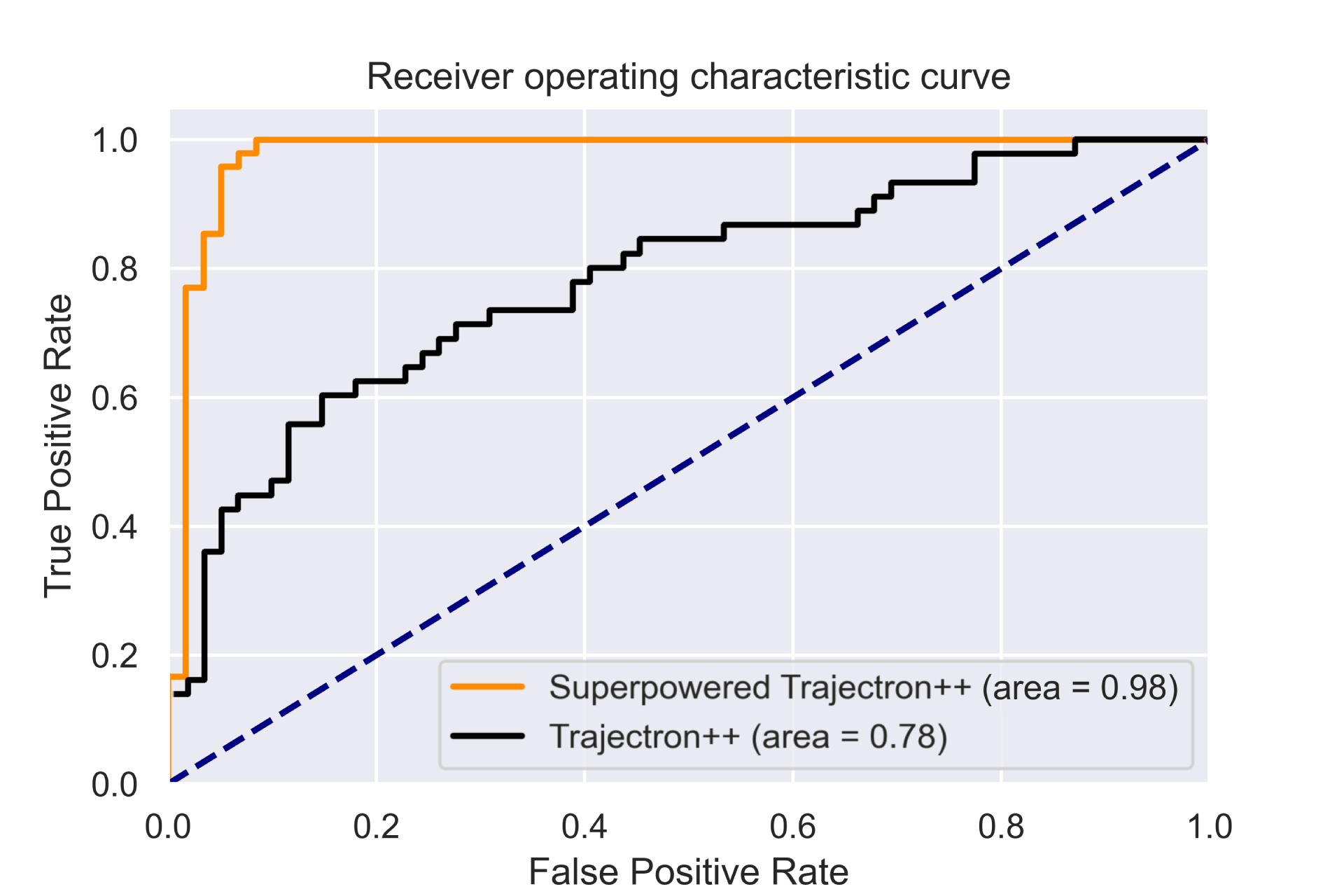}
   \caption{Trajectory Forecasting Model Comparison: The proposed superpowered trajectron++ outperforms the baseline trajectron++ model as depicted by the ROC curves. The deviations estimated from our method exhibit clear distinctions between dilemma and non-dilemma scenarios.}
   \label{fig:roc_curve}
   \vspace{-0.2in}
\end{figure}

\textbf{Effects of Dilemma Zone Features}. We evaluate the effects of the DZ features in the process of detecting dilemma zones. We use the vanilla version of Trajectron++ without DZ features to observe the effects of the DZ features. The system's performance without DZ features is compared with DZ features based on the metrics defined previously. The results are shown in Table \ref{tab:dz_effect}. The results show that the addition of DZ features improves trajectory prediction by an effective margin. The ADE and FDE are reduced by {5.15 \%} and {7.45 \%} respectively with the addition of the DZ features. 
By observing the receiver operating characteristic (ROC) curve shown in Figure \ref{fig:roc_curve}, it can be seen that the classifier works better with our model that has DZs.
The FPR is reduced greatly by {30.45\%} with the addition of DZ features. These results show that the DZ features play a significant role in detecting dilemma zones.

\begin{table}[h]
  \centering
  \begin{tabular}{@{}l|c|cccc|c@{}}
    \toprule
    \textbf{Method} & \textbf{ADE} & \multicolumn{4}{c|}{\textbf{FDE(m)}} & \textbf{FPR}\\
    & \textbf{(m)} & \textbf{@0.5s} & \textbf{@1s} & \textbf{@1.5s} & \textbf{@2s} &  \\
    \midrule
    Trajectron++ & 2.252 & \textbf{0.144} & \textbf{0.603} & 1.487 & 2.535 & 0.157\\
    Ours 		 & \textbf{2.136} & 0.183 & 0.604 & \textbf{1.401} & \textbf{2.346} & \textbf{0.109}\\
    \bottomrule
  \end{tabular}
  \caption{Evaluation of forecasting models for DZ detection}
  \label{tab:dz_effect}
  \vspace{-0.2in}
\end{table}

\begin{figure}[ht]
  \centering
	\includegraphics[width=0.8\linewidth]{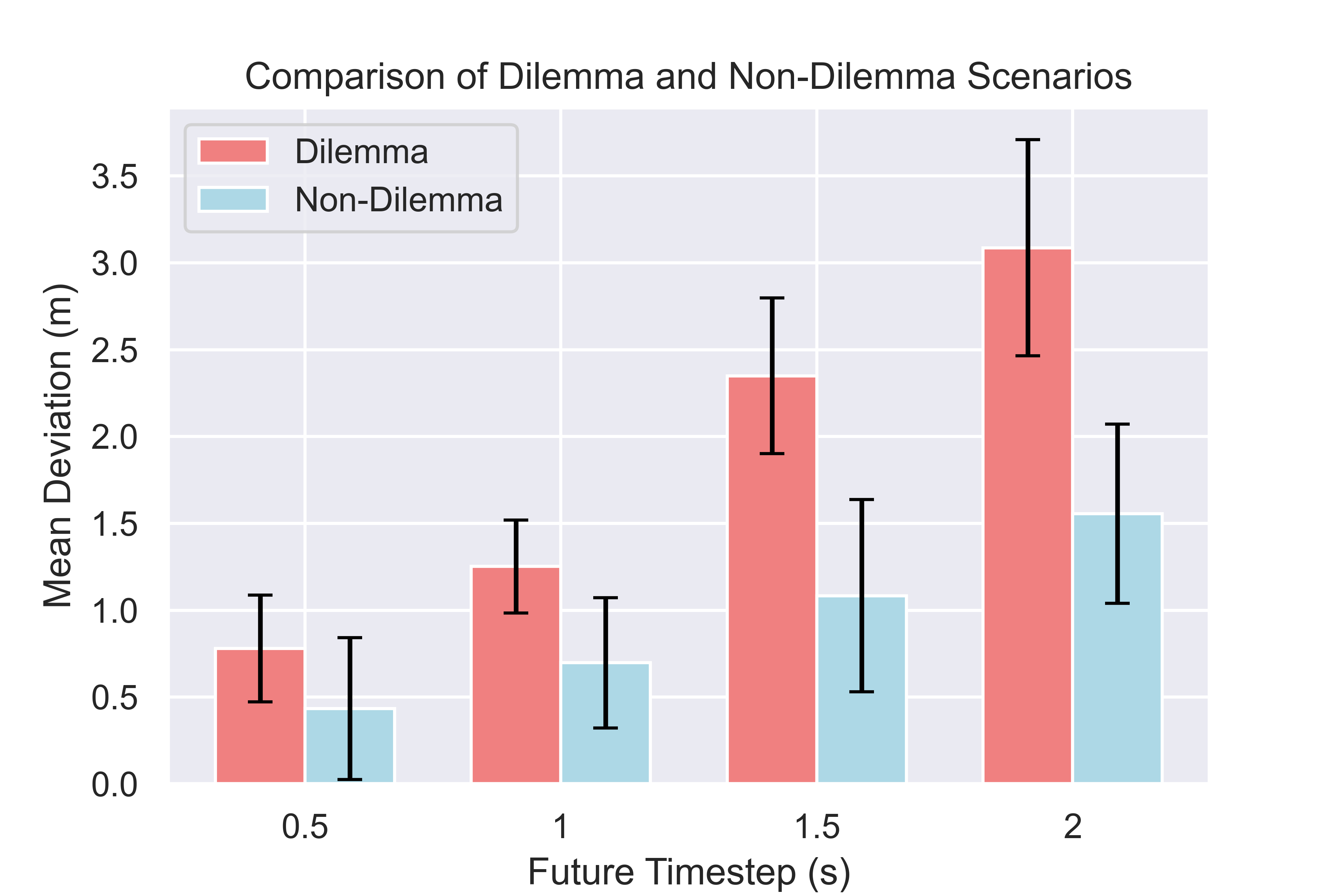}
   \caption{Dilemma v/s Non-Dilemma: A comparison of the average trajectory deviation in a yield zone}
   \label{fig:trajectory_dev}
\end{figure}

\textbf{Path Deviation under Dilemma}. 
Figure \ref{fig:trajectory_dev} shows the average trajectory deviation of vehicles inside a DZ and those that are not. It can be observed that the trajectory deviation of a vehicle inside a dilemma zone is higher than that of a non-dilemma zone. compare the average trajectory deviation of vehicles that are in a dilemma zone to those that are not.

\textbf{Maneuvering}.
The maneuver was implemented only on the ego vehicle. The acceleration of the ego was increased or decreased while the rest of the vehicles continued to follow the trajectory predicted by ST++. The simulation is run on 200 randomly sampled cases of the dilemma with 100 scenarios in which the vehicle has to accelerate and clear the intersection and vice versa. According to the US Federal Highway Authority \cite{fhwa_acc}, the maximum limit of safe acceleration is $4ms^{-2}$, which is considered the upper limit in this maneuvering experiment. The results are shown in Table \ref{tab:maneuverability}. The key observations are that vehicles need to accelerate rapidly when there is an opportunity to pass, and when the option to stop is present, deceleration should be gradual.

\begin{table}[t]
  \centering
  \begin{tabular}{@{}l|cccc@{}}
    \toprule
    Model Prediction & \multicolumn{4}{c}{\% cases without any collision for acceleration}\\
    & $2ms^{-2}$ & $4ms^{-2}$ & $-2ms^{-2}$ & $-4ms^{-2}$\\
    \midrule
    $\mathbf{P_{Pass}}>0.5$ & 83 & \textbf{94} & 3 & 21 \\
    $\mathbf{P_{Pass}} \leq 0.5$ & 5 & 6 & \textbf{91} & 79 \\
    \bottomrule
  \end{tabular}
  \caption{Ablation Study of Maneuvering Prediction}
  \label{tab:maneuverability}
  \vspace{-0.2in}
\end{table}

\section{Conclusions and Future Work}
This study introduces a novel perception-based system aimed at detecting and forecasting dilemma zones at roundabouts through trajectory prediction. By analyzing trajectory deviations in dilemma zones compared to non-dilemma zones, the system can anticipate when a vehicle is in a dilemma with a low FPR of {10\%}. However, certain limitations are acknowledged, including the system's reliance on data quality, its specific applicability to roundabouts, and potential computational challenges in real-time performance. Despite these constraints, this work lays a foundation for future advancements in traffic safety and management. We aim to extend the framework beyond roundabouts to various scenarios like unsignalized T-bone intersections and ramp-merging, enhancing real-time forecasting capabilities for intelligent transportation systems. Additionally, implement forecasting and maneuver techniques in real-world controlled scenarios to proactively address potential hazards and optimize vehicle responses and overall safety. Our team is currently working with the Institute of Automated Mobility of Arizona\footnote{https://www.azcommerce.com/iam/} and the City of Phoenix to seek real-world experiments of the presented system in the Phoenix metropolitan area. We hope this paper can collectively enhance our understanding of DZ dynamics and pave the way for more robust and proactive solutions in autonomous vehicle safety.

{\small
\bibliographystyle{IEEEtran}
\bibliography{main}
}

\end{document}